\documentclass[sigconf]{acmart}

\graphicspath{ {./images/} }
\usepackage{caption}
\usepackage{subcaption}

\AtBeginDocument{%
  \providecommand\BibTeX{{%
    \normalfont B\kern-0.5em{\scshape i\kern-0.25em b}\kern-0.8em\TeX}}}

\setcopyright{acmcopyright}
\copyrightyear{2020}
\acmYear{2020}
\acmDOI{10.1145/1122445.1122456}

\acmConference[KDD '20]{ACM SIGKDD Conference on Knowledge Discovery and Data Mining}{Aug 21--26, 2020}{San Diego, CA}
\acmBooktitle{KDD '20: ACM SIGKDD Conference on Knowledge Discovery and Data Mining,
  Aug 21--26, 2020, San Diego, CA}
\acmPrice{15.00}
\acmISBN{978-1-4503-XXXX-X/18/06}

\begin{document}

\title{Active Learning for Product Type Ontology Enhancement in E-commerce}

\author{Yun Zhu}
\affiliation{
  \institution{The Home Depot}
  \city{Atlanta}
  \state{GA}
}
\email{yun_zhu@homedepot.com}

\author{Sayyed M. Zahiri}
\affiliation{
  \institution{The Home Depot}
  \city{Atlanta}
  \state{GA}
}
\email{sayyed_m_zahiri@homedepot.com}

\author{Jiaqi Wang}
\affiliation{
  \institution{The Home Depot}
  \city{Atlanta}
  \state{GA}
}
\email{jiaqi_wang@homedepot.com}

\author{Han-Yu Chen}
\affiliation{
  \institution{The Home Depot}
  \city{Atlanta}
  \state{GA}
}
\email{hanyu_chen@homedepot.com}

\author{Faizan Javed}
\affiliation{
  \institution{The Home Depot}
  \city{Atlanta}
  \state{GA}
}
\email{faizan_javed@homedepot.com}

\renewcommand{\shortauthors}{Zhu and Zahiri, et al.}

\begin{abstract}
Entity-based semantic search has been widely adopted in modern search engines to improve search accuracy by understanding users' intent.
In e-commerce, an accurate and complete product type (PT) ontology is essential for recognizing product entities in queries and retrieving relevant products from catalog. 
However, finding product types (PTs) to construct such an ontology is usually expensive due to the considerable amount of human efforts it may involve. 
In this work, we propose an active learning framework that efficiently utilizes domain experts' knowledge for PT discovery. 
We also show the quality and coverage of the resulting PTs in the experiment results. 
\end{abstract}



\maketitle

\section{Introduction}\label{sec:intro}

\begin{figure}
     \centering
     \begin{subfigure}[b]{0.51\columnwidth}
         \centering
         \includegraphics[width=\textwidth]{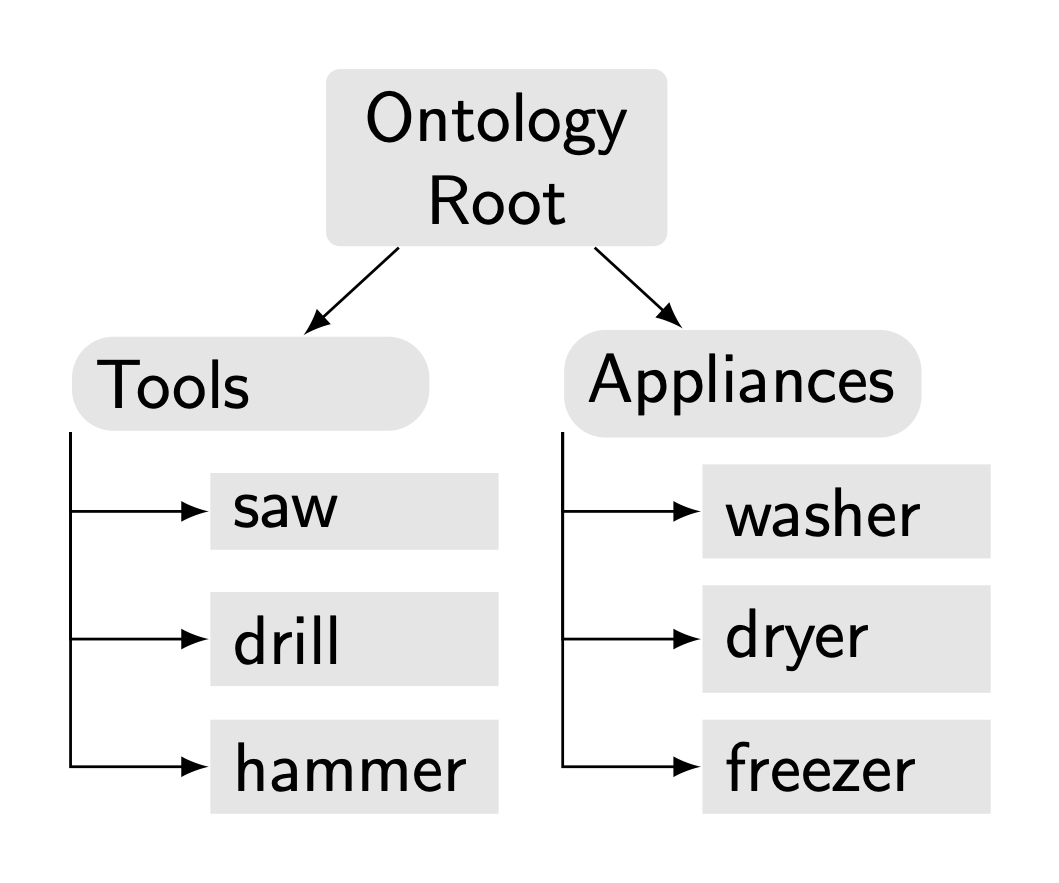}
         \caption{}
         \label{fig:ontology}
     \end{subfigure}
     \hfill
     \begin{subfigure}[b]{0.48\columnwidth}
         \centering
         \includegraphics[width=\textwidth]{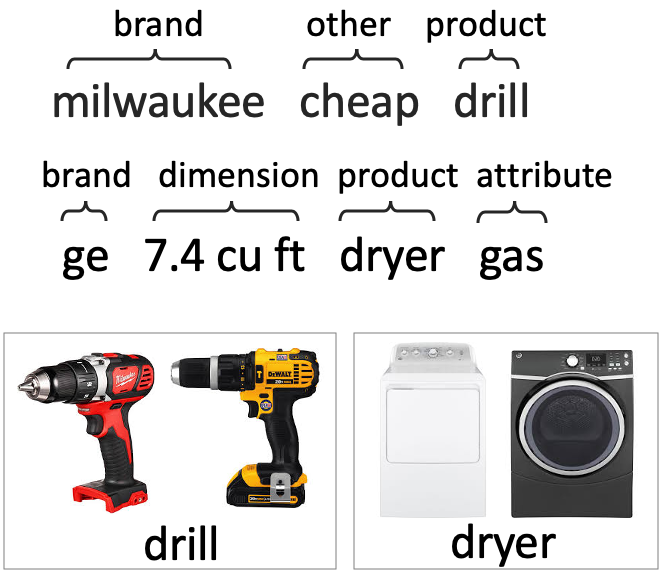}
         \caption{}
         \label{fig:ner_mapping}
     \end{subfigure}
    \caption{Example of PT Ontology (a), NER (b, top) and SKU-PT mapping (b, bottom)}
\vspace{-5mm}
\end{figure}

In the past few decades, knowledge graph construction and applications have been rapidly developed and achieved significant outcomes. 
For better relevancy in web search, Google has been leveraging knowledge graph that represents real-world entities and their relationships to one another since 2012\cite{singhal2012introducing}.
To identify those entities from text, named entity recognition (NER) techniques have been extensively studied and applied in many areas \cite{Nadeau2007NER,yadav-bethard-2018-survey} including e-commerce search \cite{Wen2019BuildingLD,Wu2017PredictingLS}.
Such NER systems usually work with a well defined ontology to classify tokens in a sequence of words \cite{Glater2017,Popov2004}.
A comprehensive and domain-specific PT ontology is beneficial to product search and discovery in an e-commerce platform \cite{Lee2006BuildingAO,kutiyanawala2018towards}.
At The Home Depot (THD), PT ontology has been used tremendously by the online search to improve query understanding and product retrieval.
For example, Figure \ref{fig:ontology} shows a snippet of our PT ontology that consists of known PT classes. The PTs in the ontology serve as the entity reference for the NER task (Figure \ref{fig:ner_mapping} top) as well as the classes for SKU-PT mapping (\ref{fig:ner_mapping} bottom) on the catalog side that facilitates the retrieval of relevant products. 
Discovering valid PTs is a key task to build or expand a PT ontology with a fundamental challenge regarding the definition of a PT. 
A PT can be defined from the demand side as atomic keywords/phrase that describes what customers look for \cite{kutiyanawala2018towards} or from the supply side as a semantic tag/label that uniquely identifies a product.
Within THD, we also have practical guidelines to distinguish between valid and invalid PTs like 
(i) no common attributes like color, brand, material, style etc in PTs (e.g., \textit{\underline{stainless steel} screw}, \textit{\underline{white} refrigerator} are not valid PTs) and
(ii) it requires significant differences in the form, functionality or usage location to make a new PT comparing to existing ones (e.g., \textit{utility sink} is qualified as it distinguishes itself from a standard \textit{sink} in its usage whereas \textit{cordless drill} is not as "cordless" doesn't change the core functionality of a \textit{drill}).

Obviously, neither the definition is definite nor the guidelines are exhaustive enough and there are always complicated cases and exceptions in which human judgement based on knowledge in merchandising, customer preference or just common sense is required.
For example, a generic PT \textit{range} can be broken down into more granular ones by fuel type like \textit{gas range}, \textit{electric range} or by other attribute like \textit{induction range}, \textit{convention range}. The word "wood" is material in \textit{wood rolling pin} while is about usage in \textit{wood glue}. 

However, leveraging human knowledge in large scale problems is usually timely and expensive. 
To reduce such cost, this paper proposes 
an active learning framework that minimizes human effort in PT discovery by 1) identifying high quality candidates using phrase mining and user behavior. 2) limiting number of PT candidates for human validation.


\section{Related Work}

Recently, incorporating structured human knowledge encapsulated in KG is proven to be very effective in various applications \cite{wang2017knowledge}. In this section we discuss some of the related works in the area of KG construction and completion. A technique to extract the information with no pre-defined ontology is proposed in \cite{lockard2019openceres}. The authors utilized semi-supervised label propagation approach to collect data and train a classifier to extract entities relations. While there are several generic knowledge bases, one of the challenges associated with domain-specific KG construction and completion is lack of publicly available knowledge base in that particular domain. To address aforementioned issue, a salable methodology is studied to expand the KG by integrating a domain-specific KG with a general domain one (such as Freebase) \cite{zhu2020collective}. The authors employed graph neural network to automatically align the entities in multiple knowledge bases. In the domain of e-commerce, several unsupervised techniques have been used to generate a commercial product-brand knowledge base by leveraging customer behavior and search terms \cite{alonso2019unsupervised}. In addition, \cite{xu2020product} provided a comprehensive comparison between generic KGs and product KGs. In this work, a self-attention based model utilized customer behavior data (queries, co-views,...) and product's content information (title, description,...) to learn product embedding and discovered the relationships between the e-commerce products. More recently, product KGs have been widely used to improve e-commerce search performance. \cite{kutiyanawala2018towards} presented unsupervised and supervised approaches to identify e-commerce product types from searched queries. They demonstrated a performance comparison between diverse approaches: (1) unsupervised product type and attribute identification directly from queries in an unsupervised fashion (2) leveraged labeled data and trained convolutional neural networks to identify product type token(s) in a query (3) trained a named entity recognition model similar to the model described by \cite{lample2016neural} to detect the product types from the queries. In this work, we introduce an active learning approach to discover new \textit{product types} by mining data from products' catalog and query logs.   

\section{method}\label{sec:method}
\begin{figure}[!htb]
        \center{\includegraphics[width=0.48\textwidth]
        {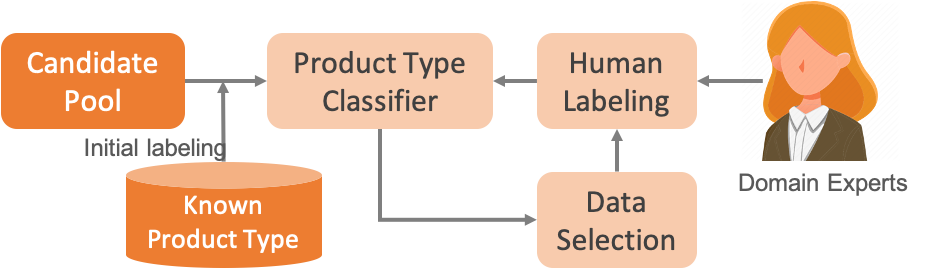}}
        \caption{\label{fig:active_learning_framework} 
        Product Type Discovery Framework}
    \end{figure}

Figure \ref{fig:active_learning_framework} shows the active learning framework of our PT discovery process that interactively involves human knowledge and machine learning techniques. The implementation of the framework is being discussed in the rest of this section.
\subsection{Candidate Pool}\label{subsec:candidate_pool}
Instead of searching among all possible words and phrases for PTs, we prepare a list of selective candidates that are more likely to be product type. These candidates are extracted from two sources: search queries and catalog content.
\subsubsection{Search Queries}
As a commonly used knowledge source for PT discovery in e-commerce \cite{kutiyanawala2018towards}, search queries draw our attention for PT candidates as customers usually specify product types explicitly or implicitly in queries with some exceptions of model number, brand, SKU number etc. In an exploratory analysis, we observe that common search queries are very likely to be PTs, e.g.,
\begin{itemize}
\item all top 10 global most frequently searched queries are PTs.
\item \textit{"flashlight"}, \textit{"flash light"}, \textit{"uv flashlight"} etc are among the top search queries for the flashlights category.
\item \textit{"ceiling fan"} and \textit{"ceiling fan with lights"} are the top 2 search queries that lead to the clicks of a particular ceiling fan SKU.
\end{itemize}
Based on this observation, we include the frequent search queries as PT candidates according to a volume threshold.
\subsubsection{Catalog Content}\label{subsubsec:catalog_content}
Apparently, high volume search queries are biased towards popular products with poor coverage of other products which haven't met the search volume threshold. Lowering the bar can help but also introduce disproportionate noisy terms with other irrelevant attributes like brand, dimension etc (e.g., "ge" and "7.4 cu ft" in the NER examples in Figure \ref{fig:ner_mapping} top). Comparing to arbitrarily formed search queries, catalog content like product title and description are in better format due to the format guidelines in our product onboarding process.
We employ the technique and tool proposed from AutoPhrase\cite{Shang2018AutoPhrase} to automatically extract quality phrases from product title and description as complementary product type candidates to common search queries.

\subsection{Known Product Type}\label{subsec:known_product_type}
Instead of building from scratch, there are thousands of known PTs previously created and validated manually.
Moreover, there are two historical versions of our PT ontology: the very first and foundation version (V1) and an expanded version (V2) developed on top of V1.
This enables the model evaluation that we can run the test on top of V1 (i.e., use V1 PTs for the initial labeling) and measure the outcome by comparing to V2 as the ground truth.
Details of evaluation are provided in Section \ref{subsec:lab_matrics}.

\subsection{Product Type Classifier}\label{subsec:product_type_classfier}
This classifier is learned from the labeled data from our domain experts and produce a confidence score of any given phrase being a valid product type. 
In each iteration of the active learning cycle, the classifier is trained using the positive-only distant training technique proposed in \cite{Shang2018AutoPhrase}  with the latest labeling by domain experts.
The implementation can be boiled down to two pieces:
\subsubsection{Training Examples}
To perform positive-only distant training, we split candidates obtained in Section \ref{subsec:candidate_pool} into a positive and a negative pool where training examples are drawn from. 
Positive pool consists of the set of valid PTs initialized by known PTs in \ref{subsec:known_product_type} and updated with new ones approved in human labeling process (described in Section \ref{subsec:human_labeling}) in each iteration. All the rest candidates form the negative pool. 
The negative pool is noisy as it contains valid product types that haven't been discovered yet. Some of these undiscovered valid PTs are being moved to positive pool if validated by domain experts. 
\subsubsection{Feature Engineering}
Given any candidate, we extract 30 features in total from the following categories: 
\begin{itemize}
\item The outcome phrase quality score from the AutoPhrase model \cite{Shang2018AutoPhrase} trained on our catalog data described in Section \ref{subsubsec:catalog_content}.
\item Intrinsic characteristics. E.g., the length, with brand name, with digits/numbers, with unit keywords like "cu ft", "mm", "volt" etc.
\item Contextual characteristics w.r.t catalog data. E.g., occurrence in product titles, position in product titles etc.
\item Contextual characteristics w.r.t search log. E.g., popularity as a search query in general and for a category and for individual SKUs, distribution of resulting clicks etc.
\end{itemize}

Following \cite{Shang2018AutoPhrase}, we train a random forest model with each base classifier an unpruned decision tree learned from a "perturbed training set" \cite{Breiman2000Randomizing}, a subset of candidates drawn with replacement from a positive and a noisy negative pool.
\subsection{Data Selection}
We don't follow the typical active learning data selection strategy that selects the most informative data points for labeling to find the optimal classification boundary \cite{ertekin2007learning} because in our scenario the classifier is cost-sensitive, i.e., not all mis-classification errors are equal.
Specifically, mistakenly approving an invalid PT could be much more damaging to our search ecosystem than missing a valid product type as many downstream applications are very sensitive the correctness of the PTs in ontology.
So we have to enforce the correctness by only approving human validated PTs. In this case, the number of new valid PTs and hence the coverage of the PT ontology is bounded by the capacity of human labeling which is usually limited.
To mitigate such limitation, we adopt a practical approach in data selection that presents domain experts the examples of high confidence score according to product type classifier for a higher yield of new valid PTs.
\subsection{Human Labeling}\label{subsec:human_labeling}
As PT inherently is a concept instead of a fact, domain experts could have different opinions in validation especially for tricky cases like the \textit{range} example mentioned in Section \ref{sec:intro}. 
To avoid such potential inconsistency and ensure the correctness, domain experts are advised to be conservative by only approving product types with great certainty and leaving others for a further review. 
This conservative strategy has an obvious impact to the classifier that the negatives are not necessarily true negatives, which echos the positive-only training technique for the product type classifier.


\section{Experiments and Results}
In this section, we report two metrics: 1) lab metrics that measure effectiveness of our method and 2) business metrics that demonstrate the impact of PTs in online search context.

Table \ref{tbl:hyperparameter} shows the hyperparameters for the product type classifier training as well as the empirically selected values by grid search used in the experiment.


\begin{table}[htbp!]
\centering\resizebox{1\columnwidth}{!}{
\begin{tabular}{p{0.42\columnwidth}|c|c}
\hline
\bf Hyperparameter & \bf Tested  & \bf Selected\\
\hline\hline
number of base classifiers   & 64, 128, 256, 512 & 256 \\
\hline
max number of features to explore at each split   & 20\%, 50\%, 80\%, 100\% & 50\% \\
\hline
number of training examples for each base classifier   & 500, 1000, 2000, 5000 & 2000 \\
\hline
\% positive training examples   & 5\%, 10\%, 20\%, 30\% & 10\% \\
\end{tabular}}
\caption{Hyperparameters Grid Search}
\label{tbl:hyperparameter}
\end{table}
\subsection{Lab Metrics}\label{subsec:lab_matrics}
Given the limited domain expertise resources, we conducted the full cycle of experiment on one category (i.e., \textit{Tools}) as the pilot study in which we discovered more than 200 new PTs.

In order to measure the effectiveness of our method in a broader range, we test it in a simulation 
by leveraging the two historical versions of the product type ontology mentioned in Section \ref{subsec:known_product_type}.
In each iteration, classifier training and data selection are performed as described in Section \ref{sec:method} but with a simulated human labeling process. Specifically, with V1 PTs as the initial positive pool, PT candidates selected according to a confidence score threshold for human labeling get approved if matching any PT in V2.
    \begin{figure}[!htb]
        \center{\includegraphics[width=0.95\columnwidth]
        {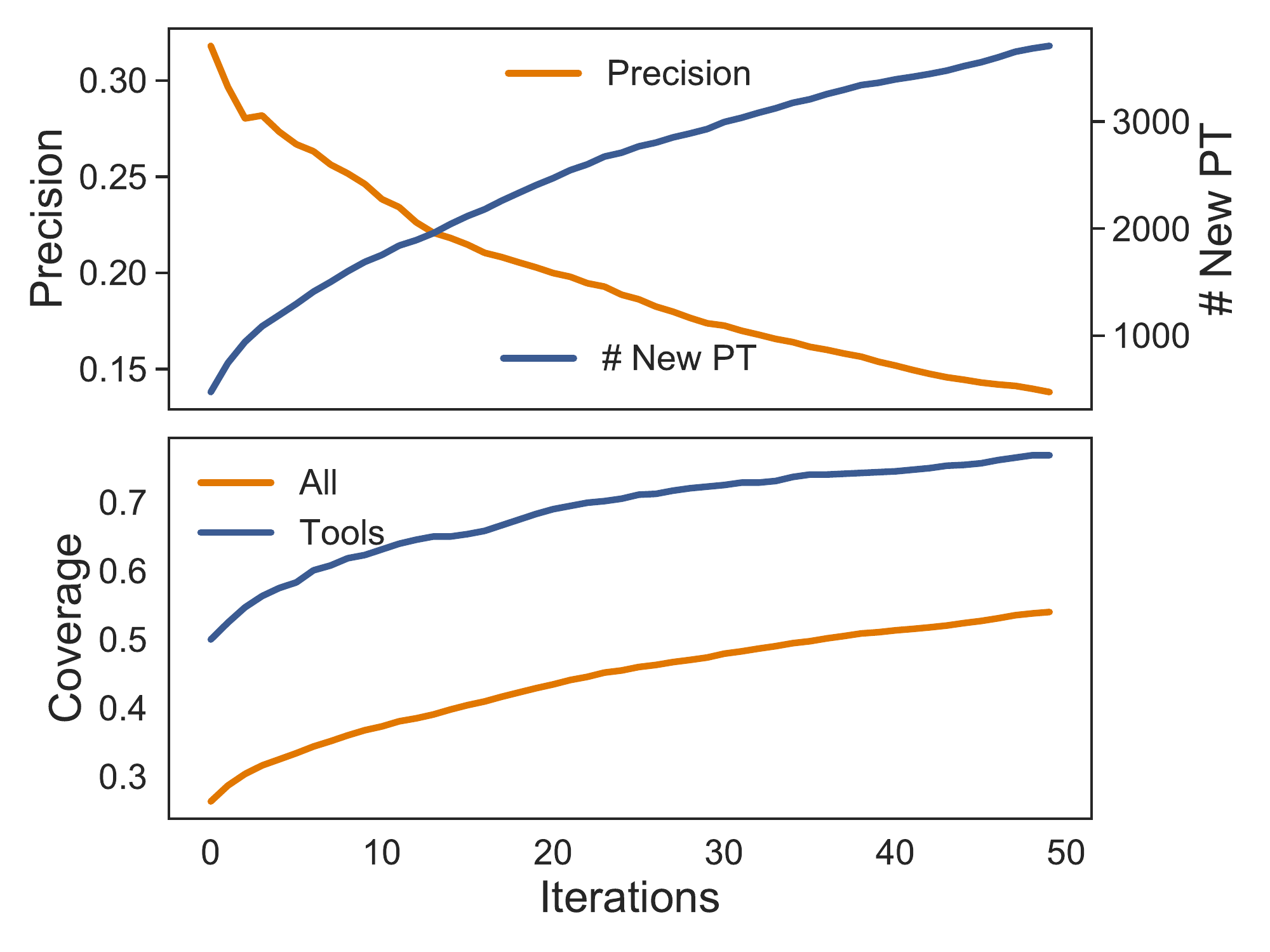}}
        \caption{\label{fig:precision_coverage} 
        Simulation Results}
    \end{figure}

\subsubsection{Effectiveness} The blue curve in Fig \ref{fig:precision_coverage} (top) shows the accumulative number of new product types being discovered as more iterations performed. More than 3500 new product types have been discovered after 50 iterations. As expected, fewer new product types can be discovered in later iterations, e.g, less than 30 in last few iterations v.s. more than 200 in the beginning. We quantified this diminishing marginal utility by precision, i.e., the ratio of correct product types among those candidates for labeling in each iteration post validations. As the orange curve indicates, precision dropped to 13\% in the last iteration from 32\% in the beginning.

\subsubsection{Coverage}
Coverage is another critical metric to measure how complete is the resulting PT ontology. i.e., if there were X product types to cover the entire catalog, how many of them are discovered.
Although the true number of product types is hard to obtain without an exhaustive labeling, we managed to estimate the coverage on the \textit{Tools} category due to the following extra efforts by our domain experts including
\begin{itemize}
    \item validate all PTs of \textit{Tools} in V2 and remove invalid or uncertain ones from the positive pool.
    \item extensively examine the \textit{Tools} category for more undiscovered PTs and add them to positive pool.
\end{itemize}
From the classifier's point of view, domain experts are essentially denoising training data for \textit{Tools} category, i.e. removing false positives and recovering missing positives.
The benefit of cleaner data is shown in Figure \ref{fig:precision_coverage} (bottom) that there is a significant lift of coverage for \textit{Tools} at over 70\% w.r.t the denoised positive PTs vs. 50\% for all categories after 50 iterations.
\subsection{Business Metric}
In online search scenario, a PT ontology provides a foundation to two key functionalities: 1) query understanding for PT recognition from query 2) PT-SKU association for relevant products retrieval.
So we measure the impact of new PTs from the following two perspectives with results shown in Table \ref{tbl:business_metrics}:
\subsubsection{PT Recognition}
High PT coverage helps to recognize PT from more queries. We sample 300k queries from one category and the percentage of queries with PT recognized is compared with and without the new PTs discovered by our model. 
\subsubsection{Search Performance}
Key search metrics including click-through rate (CTR), add-to-cart rate (ATCR) and conversion rate for a set of 150k queries sampled from another category are measured and compared for the same time period of two consecutive 
years (4th quarter of 2018 and 2019), one before and one after the new PTs are added.

\begin{table}[htbp!]
\centering\resizebox{1\columnwidth}{!}{
\begin{tabular}{c||c|c|c|c}
\hline
Business & PT &  \multicolumn{3}{c}{Search Performance}\\
\cline{3-5}
Metrics & Recognition  & CTR & ATCR & Conversion\\
\hline
\hline
Improvement   & +800 bps & +140 bps & +40 bps & +10 bps \\
\end{tabular}}
\caption{Business Metrics Summary}
\label{tbl:business_metrics}
\end{table}

\section{Conclusion \& Future Work}
In this work, we propose an active learning framework for product type discovery that leverage domain expertise in an efficient way. 
The effectiveness of the framework is demonstrated by the quality and coverage of the resulting product types in the experiments as well as the positive business impact. 
Experiment results also show that training data denoising is significantly beneficial to method performance.
There are two kinds of future work including: 1) Feature engineering of PT classifier by exploiting more textual and/or image data 2) Design a denoise procedure and add it as an additional component into the framework.

\bibliographystyle{ACM-Reference-Format}
\bibliography{sample-base}

\end{document}